# Sigma: The Key for Vision-Language-Action Models toward Telepathic Alignment

Libo Wang, Tsinghua SIGS, wanglibo@sz.tsinghua.edu.cn

***Abstract*—To address a fundamental limitation in cognitive systems, namely the absence of a time-updatable mediating thought space between semantics and continuous control, this work constructs and trains a vision–language–action model termed Sigma, deployed on a single RTX 4090. The model is built upon the open-source $\pi$0.5_base backbone, with the svla_so101_pickplace dataset preprocessed into a structured training corpus. An independently designed VLA architecture is introduced to integrate deep semantic understanding with associative reasoning, enabling telepathic-style alignment between perception and action. Training proceeds through iterative optimization of data preprocessing, LoRA-based fine-tuning, and inference-stage adapter design. Evaluation is conducted using offline closed-loop replay, comparing Sigma against the untuned $\pi$0.5_base under identical data conditions. Experimental results indicate a consistent reduction in control MSE across vector-, fragment-, and trajectory-level scales, while preserving the stability of the telepathy norm and semantic–text alignment quality. These findings demonstrate that mind-responsive alignment control can be quantitatively achieved through semantic and associative architectural integration without retraining the base model, providing a reproducible pathway for semantic alignment and intention-driven behavior.**

## I. INTRODUCTION

With the advancement of vision-language-action (VLA) systems in humanoid robotics, research has diverged into multiple parallel embodied routes rather than a single paradigm [11]. Models such as RT-2 and OpenVLA have promoted a staged pipeline that combines large-scale pre-trained vision-language models with robot-specific instructional fine-tuning [4, 13]. Typical VLA frameworks perform network-scale representation learning using visual-language backbones, integrate visual observations and linguistic commands through chain-of-thought (CoT) reasoning to form discrete latent tokens, and map them into quantized control sequences to enhance generalization and semantic reasoning [17, 20, 23]. By contrast, while adopting transformer-style architectures and multimodal tokenization, Chinese research places greater emphasis on whole-body dynamic coupling and contact stability with the physical organism [2, 28].

However, a fundamental limitation persists in current VLA architectures: the absence of a continuously updated and interpretable mediating mental space linking linguistic semantics to continuous control. This deficiency prevents the formation of stable, structured reasoning chains when the system must absorb implicit context or infer human intent. As a result, instructions with layered semantics, underspecified goals, or anthropomorphic dependencies often lead to fragmented strategies, intention drift, and semantic misalignment in humanoid robots [19]. At the same time, neither semantically driven transformer-centric pipelines nor control-oriented, hardware-coupled approaches provide an abstraction layer capable of carrying high-level semantic context while precisely aligning behavioral residuals [6]. Consequently, when a VLA system cannot internally sustain a thought space aligned with human cognitive structures, complex humanoid tasks degrade into strategic imbalance and behavioral mismatch due to disrupted idea transmission.

To bridge the gap between semantic representation and continuous control caused by the absence of a time-updatable mediating cognitive space, this study develops and releases a VLA model termed Sigma (Fig. 1). The core mechanism, telepathy, compresses deep semantic content and associative structures embedded in instructions into a continuous internal thought state, which is used to align implicit human intentions with concrete control decisions. The telepathy factor $\tau$ is formulated as a temporally shared latent cognitive vector and is integrated with dedicated perception, reasoning, and behavior modules. At the visual level, multimodal encoders with perceiver-style resampling generate visual basis tokens, which are subsequently modulated by $\tau$ through FiLM-style gating within a transformer, constraining scene representations to the current semantic and associative context.

On the language side, the multimodal large language model (MLLM) backbone first performs structured semantic factor extraction from multimodal tokens, after which these factors are progressively integrated into a temporally continuous semantic memory through the semantic workspace mechanism. By jointly incorporating contextual cues, behavioral summaries, and textual abstractions, the model infers latent intentions that are not explicitly expressed in the input. These inferred intentions are then projected into the telepathy factor $\tau$ by the telepathy projector, yielding a shared internal thinking workspace that maintains semantic coherence over time and supports deep, association-level understanding across perception, reasoning, and action.

For the action module, Sigma first computes a safe baseline behavior under the condition $\tau = 0$, establishing a stable reference policy. It then combines $\tau$ with higher-order representations to generate an explicit residual $\Delta a$ via a telepathy residual head. After residual fusion and translation into motor commands by the low-level controller, the module enables behavior modulation that closely approximates mind-ready human intention while maintaining physical stability.

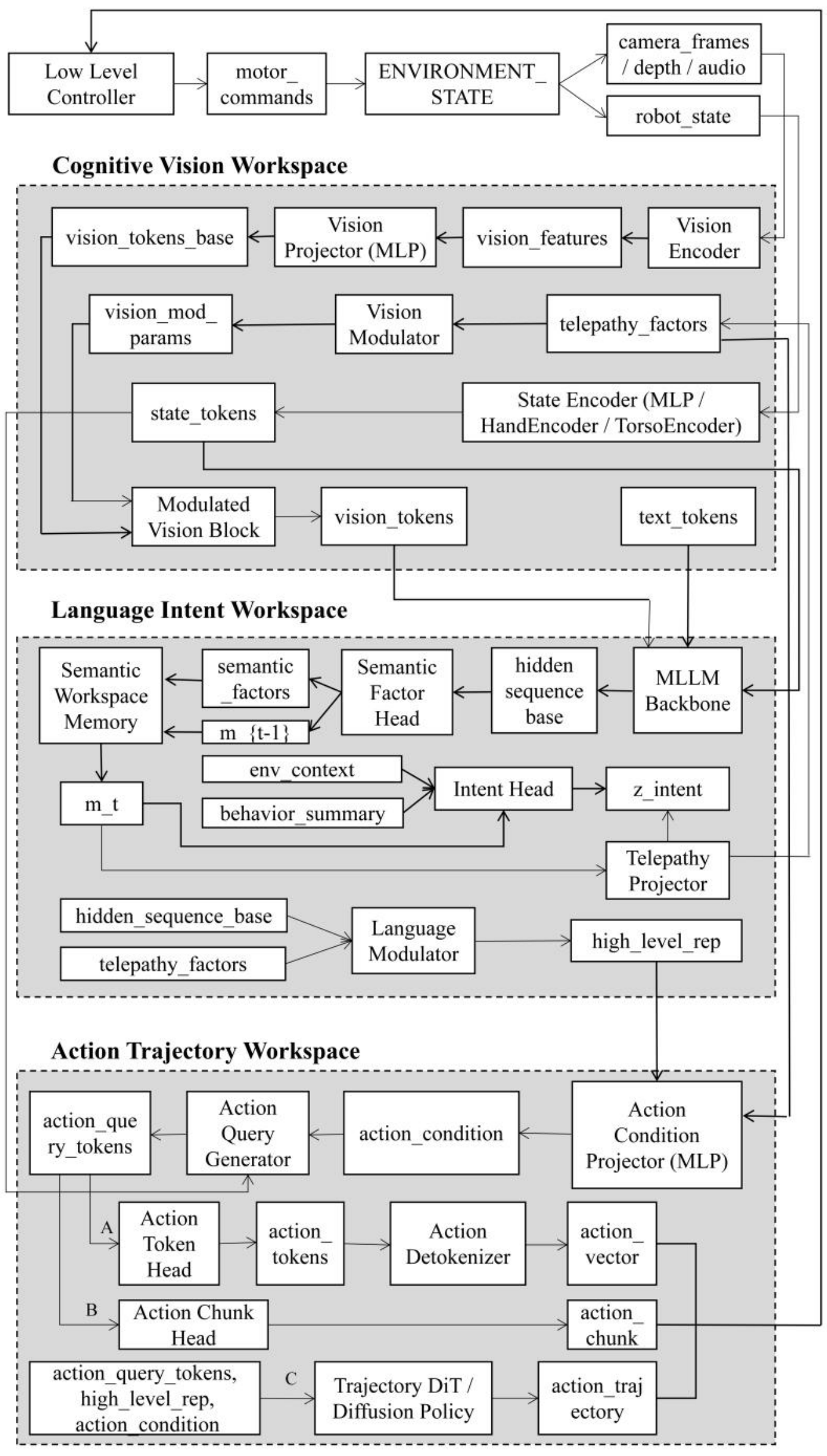

Fig. 1 Architecture of Sigma

## II. RELATED WORK

### A. Vision-Language-Action Models

Architecturally, vision language action models act as multimodal foundation policies that couple visual perception, language conditioning, and action generation [14, 24]. Most pipelines treat a pretrained vision language model as the semantic backbone, encoding observations and commands into latents that an action decoder maps to joint space trajectories or discrete control symbols executable on hardware [19, 22, 25]. Training commonly follows network scale image and text pretraining and robot trajectory fine tuning on aligned triplets [5, 12]. RT 2 tokenizes actions as text, whereas OpenVLA and the $\pi$0.5 series learn transferable control from cross platform corpora such as Open X Embodiments [1, 4, 13]. Newer $\pi$0.5 and LeVERB adopt layered control to separate high level semantics from high frequency actuation, balancing generalization and motion fidelity [7, 10].

### B. Multimodal Learning

Within a unified paradigm, multimodal learning addresses the joint processing of heterogeneous signals, including language, vision, audio, and motion, by mapping formerly isolated perceptual channels into a shared latent representation space [3, 29]. Dedicated modality-specific encoders compress images, text, and state trajectories into aligned vectors, while contrastive or joint embedding objectives enforce cross-modal semantic consistency and suppress irrelevant samples, yielding stable correspondences [9, 16]. With the widespread adoption of transformer architectures, cross-modal attention and iterative interaction layers have become dominant, enabling simultaneous focus on textual elements and visual regions and progressively refining their associations. This mechanism directly supports downstream alignment, retrieval, and decision control in embodied systems [2, 8, 15, 25, 26].

## III. ARCHITECTURE

Unlike prior thought-communication schemes that bypass natural language through multi-agent exchange, this architecture internalizes latent reasoning as a shared telepathy–semantic memory space within a single humanoid, enabling residual-based modulation of baseline control as intrinsic perception-action alignment.

### A. Abbreviations and Acronyms

The module begins with environmental feedback driven by motor commands from the low-level controller. Sensor outputs including camera frames, depth, audio, and robot state are routed to the vision and state encoders. PatchEmbed Conv and AudioPatchEmbed transform these signals into variable-length tokens that are aggregated as vision features. A two-layer MLP vision projector maps them into the MLLM d_model space. A perceiver-style resampler then uses learnable queries to produce fixed-length vision tokens, ensuring a stable visual basis for the language module.

The state encoder compresses robot state information through an MLP and expands it into $N_s$ state tokens via a token expander, aligning proprioceptive and visual signals within a shared representation space. Telepathy factors are scaled by a learnable $\tau_{\text{log_scale}}$ and injected into the vision modulator to produce FiLM parameters $\gamma$ and $\beta$, enabling channel-wise modulation of the vision token base. A modulation scale further allows smooth interpolation between neutral perception and latent thought emphasis. In parallel, textual commands are tokenized externally into text tokens that share the same d_model space, enabling unified cross-modal reasoning with vision, state, and telepathy signals in the language module.

The modulated vision block refines local relations on v_mod through multi layer self attention and feedforward networks, producing semantically saturated vision tokens shaped by telepathy. Together with state_tokens, they enter the language intent workspace as the perceptual input of the mind sensing chain.

$$F \in \Re^{N_f \times d},\ Q \in \Re^{N_f \times d},\ \tau \in \Re^{d_\tau}$$

$$A = \mathrm{softmax}\left(\frac{QW_Q (FW_K)^{\mathrm{T}}}{\sqrt{d}}\right),\ H = A(FW_V)$$

$$V_{\text{base}} = \mathrm{FFN}\,(\mathrm{LN}(\mathrm{Q}+\mathrm{H})),\ V_{\text{base}} \in \Re^{N_v \times d}$$

$$\tau' = e^{\theta_\tau} \tau,\ h = \sigma(\tau' W_1 + b_1),\ [\gamma, \beta] = hW_2 + b_2$$

where $F$ is the multimodal visual feature sequence output by the vision encoder and projector, $N_f$ is the number of visual tokens in the sequence; $Q$ is the query token sequence used for resampling by the perceiver; $N_v$ is the fixed length of the output visual tokens; d is the shared feature dimension dmodel; $W_Q$, $W_K$, and $W_V$ are the linear projection matrices of query, key, and value, respectively; $A$ is the attention weight matrix calculated; $H$ is the feature vector aggregated according to $A$; LN represents the LayerNorm applied to the token sequence; FFN represents the feedforward sublayer; $V_{\text{base}}$ is the fixed-length visual base token sequence obtained after resampling; $\tau$ represents the telepathy factor vector from the language module; $d_\tau$ is its dimension; $\theta_\tau$ is the log-scale gate parameter controlling the telepathy intensity; and $\tau'$ is the factor amplified by $\theta_\tau$. $W_1$, $W_2$ and $b_1$, $b_2$ are the weights and biases of the MLP inside the vision modulator; $\sigma(\cdot)$ is the GELU nonlinearity used therein; $\gamma$ and $\beta$ are the FiLM scaling and translation coefficients generated by $\tau'$; $V_{\text{base}}$ is the unmodulated visual base token; $V_{\text{film}}$ is the intermediate visual representation after applying FiLM; $\theta_{\text{mod}}$ is the second log-scale gate parameter controlling the modulation amplitude; and $V_{\text{mod}}$ is the final output telepathy modulated visual token.

### B. *Language Intent Workspace*

After establishing a high level linguistic thought field across time, semantics, and context, this module concatenates text, vision, and state tokens as its unified input. An MLLM backbone applies multi layer cross modal attention to construct a hidden sequence that aligns language, perception, and ontological state within the shared d_model space. A semantic factor head extracts $K$ semantic factors from this sequence using learnable queries. At each time step, the semantic workspace memory retrieves the previous semantic state $m_{t-1}$ and integrates current factors into $m_t$ through a gated recursive update, ensuring temporal semantic continuity.

Three summary heads independently extract env_context, behavior_summary, and text_summary from the hidden sequence base, allowing explicit separation of environmental cues, behavioral patterns, and linguistic context. The intent head integrates $m_t$, $c_{\text{env}}$, and $c_{\beta\text{eh}}$ to infer $z_{\text{intent}}$, which functions as the semantic driver of telepathy. The telepathy projector then fuses $m_t$, $z_{\text{intent}}$, $c_{\text{env}}$, $c_{\beta\text{eh}}$, $z_{\text{sem-pool}}$, and $c_{\text{text}}$ to produce telepathy factors as a global semantic alignment vector. Guided by $\tau_t$, the language modulator applies gated bias modulation to the hidden sequence, yielding a high level semantic representation that feeds back into other modules to complete the cross modal closed loop.

$$Z_{\text{sem}} \in \Re^{K \times d},\ Z_{\text{pool}} = \frac{1}{K}\sum_{i=1}^{K} Z_{\text{sem},i}$$

$$u = \mathrm{GELU}\ (W_u z_{\text{pool}} + b_u),\ \lambda = \sigma(W_\lambda [m_{t-1}; z_{\text{pool}}] + b_\lambda),\ m_t = \lambda \odot m_{t-1} + (1-\lambda) \odot u$$

$$x = [m_t; z_{\text{intent}}; c_{\text{env}}; c_{\text{beh}}; z_{\text{sem_pool}}; c_{\text{text}}]$$

$$h = \mathrm{GELU}(W_1 x + b_1),\ \tau_t = W_2 h + b_2$$

Where $m_{t-1}$ represents the semantic memory vector at the previous time step in semantic memory propagation; $z_{\text{sem}}$ is the semantic factor matrix read from the semantic factor head at the current time step; $z_{\text{pool}}$ is the average pooling result of $z_{\text{sem}}$; $u$ is the candidate semantic signal after updated projection; $\lambda$ is the gating coefficient for interpolation between the old memory $m_{t-1}$ and the candidate update $u$; $m_t$ is the current semantic memory vector obtained after integration; in the telepathy projector, mt serves as the semantic memory summary; $z_{\text{intent}}$ is the latent intent vector inferred from the intent head; $c_{\text{env}}$ and $c_{\text{beh}}$ correspond to contextual summarization and behavioral trend summarization, respectively; $z_{\text{sem_pool}}$ is the pooling representation of semantic factors at each time step; $c_{\text{text}}$ is the text summary. These six vectors are concatenated to form a fusion vector $x$, $h$ represents the implicit semantics of $x$ after transformation by a multi-layer perceptron; and $\tau_t$ represents the final projected telepathy_factors used to modulate higher-order semantic alignment in other modules.

### C. *Action Trajectory Workspace*

This module converts high_level_rep and the current perceptual state into executable control trajectories. The action condition projector merges high_level_rep with telepathy_factors to generate action_conditions via an MLP for planning. The action query generator integrates action_conditions with state_tokens, applies cross attention and transformer refinement on learnable queries, and outputs action_query_tokens as a shared basis for action branches.

Along path A, the action token head compresses action query tokens into low dimensional tokens, which the action tokenizer reduces to a continuous action vector for single step or high frequency control. Along path B, the action chunk head uses pooling and linear projection to form short, coherent action segments. Along path C, a trajectory diffusion policy conditions on action queries, high level representations, and action conditions to denoise trajectories into longer horizon plans. The action vector, chunks, and trajectory are reweighted and decoded by action fusion and the low level controller into motor commands, completing the intent to execution loop.

$$Q_1 = \mathrm{Attn}(Q_0, S_{\text{proj}}, S_{\text{proj}}) + b(c_{\text{act}}),\ q_t = \mathrm{LN}\,(\mathrm{Ref}(Q_1))$$

$$c_{\text{act}}^{\text{base}} = P_{\text{act}}(r_{\text{high}}, 0),\ q_{\text{base}} = Q\,(c_{\text{act}}^{\text{base}}, S_t)$$

$$a_{\text{vec}}^{\text{base}} = H_A(q_{\text{base}}),\ a_{\text{chunk}}^{\text{base}} = H_B(q_{\text{base}}),\ a_{\text{traj}}^{\text{base}} = D_{traj}(q_{\text{base}}, r_{\text{high}}, c_{\text{act}}^{\text{base}})$$

$$[\Delta a_{\text{vec}}, \Delta a_{\text{chunk}}, \Delta a_{\text{traj}}] = \mathrm{g}([r_{\text{high}}, \tau])$$

$$a_{\text{vec}}^{\tau} = a_{\text{vec}}^{\text{base}} + \Delta a_{\text{vec}},\ a_{\text{chunk}}^{\tau} = a_{\text{chunk}}^{\text{base}} + \Delta a_{\text{chunk}},\ a_{\text{traj}}^{\tau} = a_{\text{traj}}^{\text{base}} + \Delta a_{\text{traj}}$$

$$u_t = \phi\,(a_{\text{vec}}^{\tau}, a_{\text{chunk}}^{\tau}, a_{\text{traj}}^{\tau}),\ m_t = C_{\text{low}}(u_t)$$

where $S_t$ represents state_tokens; $W_s$ is the linear projection matrix that aligns them to d_model; $Q_{\text{seed}}$ is the learnable query template; $\mathrm{Attn}(\cdot)$ represents multi-head cross-attention operation; $c_{\text{act}}$ is the condition vector generated by the action condition projector; $b(\cdot)$ is the bias it applies to the query; Ref $(\cdot)$ is the transformer encoder on the query; $\mathrm{LN}(\cdot)$ is layer normalization; $q_t$ is action_query_tokens; $P_{\text{act}}$ represents the action condition projector; $Q$ represents the action query generator; $H_A$ and $H_B$ correspond to the action token head and action chunk head, respectively; $D_{\text{traj}}$ corresponds to the trajectory DiT/diffusion policy; $a^{\text{base}}{}_{\text{vec}}$, $a^{\text{base}}{}_{\text{chunk}}$, and $a^{\text{base}}{}_{\text{traj}}$ are the three-way baseline actions; $g(\cdot)$ is the MLP of the telepathy residual head; $\Delta a_{\text{vec}}$, $\Delta a_{\text{chunk}}$, and $\Delta a_{\text{traj}}$ are the telepathy residuals, $a^{\tau}_{\text{vec}}$, $a^{\tau}_{\text{chunk}}$ and $a^{\tau}_{\text{traj}}$ are the final action branches; $\phi(\cdot)$ is action fusion; $u_t$ is the control representation after fusion; $C_{\text{low}}$ is the low-level controller; $m_t$ corresponds to the motor_commands output after flowing to the low-level controller in the graph.

## IV. TRAIN

A complete and transparent training pipeline is adopted. Multimodal sequence data are first preprocessed in PyTorch to unify visual, linguistic, and proprioceptive signals within a single representation. LoRA fine-tuning is then applied to the open source VLA model $\pi$0.5_base on a single RTX 4090 to stabilize semantic to action mapping. Hardware performance is reported in Fig. 2.

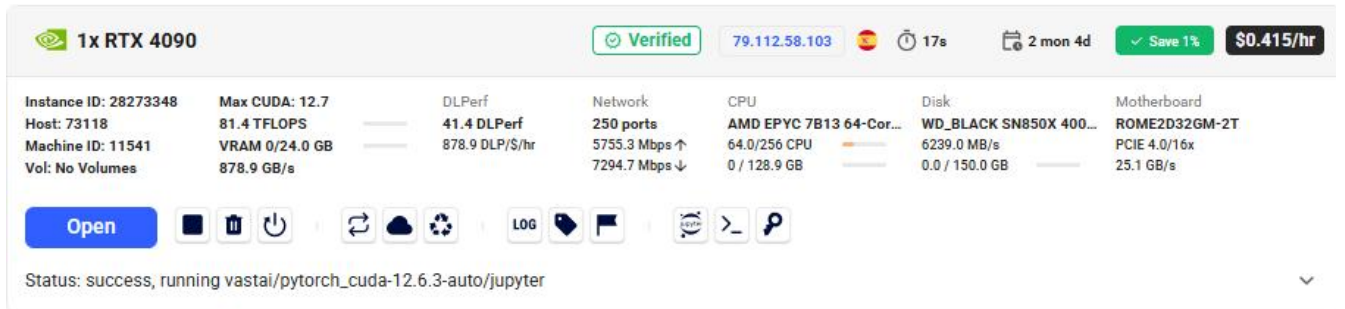

Fig. 2: Detailed computing power configuration of a single NVIDIA RTX 4090

In addition, an interventional adapter is integrated to enhance controllability and fine-grained semantic correction during inference. This mechanism enables semantic intervention without modifying the underlying model parameters.

### A. Data Preprocessing

The preprocessing pipeline is fully automated via scripts. load_sigma_env first loads environment variables and HF_token from sigma.env, followed by load_lerobot_dataset to ingest svla_so101_pickplace trajectories. To stabilize large downloads, prefetch_hf_dataset warms the Hugging Face cache with exponential backoff and 429 handling. Data iteration uses safe_iter_dataset with retry logic, while samples are grouped by episode_index and temporally ordered by frame_index when required.

For each episode, trajectories are segmented using a sliding window with horizon_$T$ = 16 via build_windows. Within each window, dedicated extractors retrieve visual frames, robot state, action sequences, and text commands, after which compute_action_stats derives the average and maximum L2 action norms. Windows with average norms below the min_action_norm threshold are treated as near-static and removed to retain only effective operations. For each retained window, a standardized sample dictionary is constructed, mapping raw modalities to training-ready fields including vision_inputs, robot_state, ground-truth action vectors, chunks, trajectories, and associated norm statistics.

Sample writing is managed by ShardWriter, which scans existing shard_*.pt files at initialization to determine the starting shard index and skip_count, preventing duplicate writes. Data are sequentially emitted as shard_00000.pt, shard_00001.pt, shard_00002.pt after filling a fixed shard_size buffer. In parallel, meta.json records dataset IDs, episode counts, window statistics, and preprocessing hyperparameters to ensure traceable fine-tuning and evaluation.

### B. LoRA Fine-Tuning

As the LoRA fine-tuning stage of Sigma, the training procedure is implemented through a fully specified engineering pipeline. The process begins by loading the visual-language backbone $\pi$0.5_base from Hugging Face or a local cache [10]. A LoRA configuration with $r$ = 16 and dropout is applied, unfreezing only the $q$, $k$, $v$, and output projection layers, while four-bit quantization, mixed precision, and gradient accumulation are enabled to control computational overhead. The Sigma shard dataset is then constructed from the generated shard files. A custom collator encodes text commands into one-hot vectors and projects them into d_model space. In parallel, multi-frame visual features and compressed robot_state are unified into vision_inputs and state tensors, and vector-, chunk-, and trajectory-level ground truth actions, together with optional baselines, are derived to support residual learning.

The proposed architecture integrates vision, language, and action submodules within a unified forward pass. Each step first executes visual inference without telepathy, after which the language module generates telepathy_factors and high_level_rep that are fed back to produce three action outputs. The following algorithms govern fine-tuning behavior.

### C. Telepathic Residual Action Focusing

The Telepathic Residual Action Focusing (TRAF) algorithm uses the $\pi$0.5 baseline as a reference and learns only telepathic residuals to inject high-level semantics into action_vector, action_chunk, and action_trajectory. It further computes sample-wise errors and upweights top-$k$ difficult segments, concentrating learning on challenging alignment cases while preserving control stability.

For each batch, the model outputs three raw actions;

$$\Delta a_{\text{vec}}, \Delta a_{\text{chk}}, \Delta a_{\text{traj}}$$

If the data also provides the baseline ($\pi$0.5) action abase:

$$a_{\text{vec}} = a_{\text{vec}}^{\text{base}} + \Delta a_{\text{vec}},\ a_{\text{chk}} = a_{\text{chk}}^{\text{base}} + \Delta a_{\text{chk}},\ a_{\text{traj}} = a_{\text{traj}}^{\text{base}} + \Delta a_{\text{traj}}$$

Otherwise, $a = \Delta a$, the corresponding annotation is:

$$a^{*}_{\text{vec}}, a^{*}_{\text{chk}}, a^{*}_{\text{traj}}$$

The global action loss is:

$$L_{\text{act}} = \alpha_a \text{MSE}(a_{\text{vec}}, a^{*}_{\text{vec}}) + \alpha_b \text{MSE}(a_{\text{chk}}, a^{*}_{\text{chk}}) + \alpha_c \text{MSE}(a_{\text{traj}}, a^{*}_{\text{traj}})$$

Calculate the difficulty of each sample based on the telepathy-corrected final action:

$$h_i = \text{MSE}_i(a_{\text{vec}}, a^{*}_{\text{vec}}) + \text{MSE}_i(a_{\text{chk}}, a^{*}_{\text{chk}}) + \text{MSE}_i(a_{\text{traj}}, a^{*}_{\text{traj}})$$

Select the top-$k$ subsets (with a proportion of $\rho$) to form set H, and then calculate a difficult sample loss on this subset:

$$L_{\text{hard}}^{\text{act}} = \alpha_a \text{MSE}_H(a_{\text{vec}}, a^{*}_{\text{vec}}) + \alpha_b \text{MSE}_H(a_{\text{chk}}, a^{*}_{\text{chk}}) + \alpha_c \text{MSE}_H(a_{\text{traj}}, a^{*}_{\text{traj}})$$

The final action item is

$$L_{\text{act,total}} = L_{\text{act}} + \lambda_{\text{hard}} L_{\text{act}}^{\text{hard}}$$

where $a_{\text{vec}}$, $a_{\text{chk}}$, and $a_{\text{traj}}$ are the final vector-level, fragment-level, and trajectory-level action outputs; $a_{\text{chk}}$ is the corresponding label; $\Delta a.$ is the telepathy residual head prediction; $a_{.}^{\text{base}}$ is the offline $\pi$0.5 baseline action; $\alpha_a$, $\alpha_b$, and $\alpha_c$ are the loss weights of the three actions; $h_i$ is the difficulty score of the i-th sample; $p$ is the proportion of difficult samples; $H$ is the set of top-k difficult samples; $\lambda_{\text{hard}}$ is the loss weight of difficult samples.

## D. Telepathic Residual Action Focusing

The Telepathic Semantic Alignment Curriculum (TSAC) progressively modulates the alignment weights among semantic memory, intention vectors, and telepathy factors throughout training. The curriculum first emphasizes action regression to stabilize fundamental control behavior, and subsequently increases semantic consistency and directional regularity in a linear schedule. This staged adjustment guides the model to gradually align its internal thought structure with explicit action trajectories, leading both representations to converge toward a shared mental coordinate system in the later training phases. The corresponding algorithm is detailed as follows:

Semantic consistency loss is composed of the semantic factor pooling vector $z_{\text{pool}}$, the previous moment's memory $m_{\text{prev}}$, and the text/vision pooling vector.

$$L_{\text{sem}} = L_{\text{time}}(z_{\text{pool}}, m_{\text{prev}}) + \beta_{\text{mi}} L_{\text{mi}}(z_{\text{pool}}, \text{text} + \text{vision})$$

Intent association loss involves aligning the intent vector $z_{\text{intent}}$ with the current semantic memory $m_t$ in terms of direction:

$$L_\tau = 0.01 \| \tau \|_2^2 + \lambda_{\text{collapse}} [\max(0, \tau_0 - \| \tau \|_2)]^2 + \eta_{\text{var}} (1 - \cos(\tau, c_{\text{act}}))$$

Overall loss per training step t using course weighting:

$$w_{\text{sem}}(t),\ w_{\text{intent}}(t),\ w_\tau(t)$$

Linear interpolation from the initial value to the target value yields:

$$L_{\text{total}}(t) = L_{\text{act,total}} + w_{\text{sem}}(t) L_{\text{sem}} + w_{\text{intent}}(t) L_{\text{intent}} + w_\tau(t) L_\tau$$

where $z_{\text{pool}}$ is the semantic factor pooling vector; mprev and mt are the semantic memories of the previous and current moments, respectively; $L_{\text{time}}$ measures the consistency of semantic memory across time; $L_{\text{mi}}$ is the mutual information comparison loss between the semantic vector and the text/vision fusion representation; $\beta_{\text{mi}}$ is its weight; $z_{\text{intent}}$ is the intention vector; $\tau$ is the telepathy_factors; $c_{\text{act}}$ is the action_condition; $\lambda_{\text{collapse}}$ controls the strength of the anti-collapse term; $\tau_0$ is the minimum norm threshold of the expectation; $\eta_{\text{var}}$ is the direction alignment regularization weight; $w_{\text{sem}}(t)$, $w_{\text{intent}}(t)$, and $w_\tau(t)$ are the course weights that increase linearly with the number of training steps, making the early stage mainly action regression; the late stage then gradually strengthens semantic and telepathy alignment.

## E. Adapter

As an auxiliary mechanism for optimizing LoRA performance, an inference-stage adapter is introduced to intervene in Sigma's behavioral outputs through a hook-and-loop control scheme, without modifying the underlying model weights [18]. The adapter simultaneously reads the model-predicted action_vector, action_chunk, and action_trajectory, together with the offline base_action produced by $\pi$0.5_base, and defines their difference as the telepathic residual $\Delta a_\tau$. A risk score is then computed by jointly considering the residual-to-baseline norm ratio of $\Delta a_\tau$, the L2 norm of telepathy_factors, and the cosine similarity between telepathy_factors and action_condition. This score is mapped through an exponential function to obtain a continuous scaling factor bounded between minscale and maxscale. When the residual magnitude remains moderate and $\tau$ is well aligned with the action conditions, the adapter amplifies telepathic residual weights, enabling effective high-level semantic correction. Conversely, when residuals become unstable or $\tau$ deviates from the target range, the gating mechanism smoothly suppresses the residual, reverting the output toward the $\pi$0.5 baseline behavior. The final outputs include the adapted action_vector, action_chunk, and action_trajectory, together with the corresponding scaling factors and risk metrics, thereby preserving intuitive telepathic adjustments while enforcing fine-grained risk control to maintain overall control stability.

## F. Loss Dynamics

Given the established LoRA and adapter pipeline, the training loss dynamics were systematically analyzed. As reported in Table 1, measurements from epoch 0 indicate that the total loss is dominated by $L_{\text{actt}}$, while $L_{\text{sem}}$ remains stable at approximately 0.07. The intention loss $L_{\text{int}}$ gradually shifts from near zero to a clear negative range, reflecting an increasing cosine similarity between intention vectors and semantic memory as curriculum weights rise, thereby avoiding semantic collapse. As $w_{\text{sem}}$, $w_{\text{int}}$, and $w_{\text{tau}}$ increase linearly, $\tau_{\text{rms}}$ smoothly grows from about 0.04 to 5.6, indicating controlled activation of telepathy magnitude, while $L_{\text{tau}}$ decreases to roughly 0.06–0.33. Together with a stable hard_ratio = 0.30 and comparable fluctuations in $L_{\text{act_hard}}$, no gradient explosion or mode collapse is observed.

Table 1 shows that action regression, semantic alignment, and telepathy regularization remain balanced. For Sigma, baseline control stability is preserved while representational capacity shifts toward semantic–intention–action alignment, yielding measurable telepathic advantage.

# V. EXPERIMENTS

Offline closed-loop replays were performed to compare the control behavior of Sigma and $\pi$0.5_base on identical datasets. Using shard-generated trajectories, success rates and trajectory errors were statistically evaluated with telepathy enabled and disabled to quantify the contribution of telepathic alignment. The design isolates the effect of telepathic modulation from data and policy variability, enabling a controlled assessment of its behavioral impact.

TABLE 1: Decomposed training loss and telepathy activation profile

| step/gstep | loss | L_act | L_sem | L_int | L_tau | L_act_hard | w_sem | w_int | w_tau | hard_ratio | tau_rms |
|---|---|---|---|---|---|---|---|---|---|---|---|
| 0 | 1086.908 | 1086.898 | 0.069 | 0.032 | 0.172 | 618.710 | 0.100 | 0.100 | 0.000 | 0.300 | 0.049 |
| 10 | 1914.058 | 1914.036 | 0.069 | 0.035 | 0.171 | 1141.169 | 0.210 | 0.185 | 0.004 | 0.300 | 0.049 |
| 20 | 1308.330 | 1308.308 | 0.069 | -0.004 | 0.169 | 716.477 | 0.320 | 0.271 | 0.007 | 0.300 | 0.044 |
| 30 | 1033.426 | 1033.469 | 0.069 | -0.209 | 0.156 | 573.119 | 0.429 | 0.356 | 0.011 | 0.300 | 0.054 |
| 40 | 1838.295 | 1838.500 | 0.073 | -0.558 | 0.112 | 1100.001 | 0.539 | 0.441 | 0.015 | 0.300 | 0.146 |
| 50 | 1103.291 | 1103.584 | 0.069 | -0.645 | 0.090 | 580.732 | 0.649 | 0.527 | 0.018 | 0.300 | 0.241 |
| 60 | 1426.918 | 1427.275 | 0.069 | -0.671 | 0.072 | 748.638 | 0.759 | 0.612 | 0.022 | 0.300 | 0.381 |
| 70 | 2036.956 | 2037.376 | 0.069 | -0.691 | 0.058 | 1181.513 | 0.868 | 0.698 | 0.026 | 0.300 | 0.566 |
| 80 | 2080.200 | 2080.660 | 0.118 | -0.737 | 0.045 | 1148.737 | 0.978 | 0.783 | 0.029 | 0.300 | 1.075 |
| 90 | 2319.583 | 2320.092 | 0.069 | -0.725 | 0.049 | 1162.648 | 1.000 | 0.800 | 0.030 | 0.300 | 1.435 |
| 100 | 1975.697 | 1976.183 | 0.069 | -0.696 | 0.061 | 1133.261 | 1.000 | 0.800 | 0.030 | 0.300 | 1.898 |
| 110 | 1112.991 | 1113.447 | 0.069 | -0.659 | 0.085 | 585.033 | 1.000 | 0.800 | 0.030 | 0.300 | 2.503 |
| 120 | 1058.931 | 1058.888 | 0.525 | -0.610 | 0.202 | 535.251 | 1.000 | 0.800 | 0.030 | 0.300 | 4.292 |
| 130 | 1669.481 | 1669.870 | 0.069 | -0.586 | 0.337 | 835.426 | 1.000 | 0.800 | 0.030 | 0.300 | 5.653 |

### *A. Setup*

Under a fixed pipeline, the experiment isolates the effect of the telepathy layer by toggling its activation within the same backbone. The experimental condition uses Sigma, a $\pi$0.5_base model with LoRA fine-tuning and adaptation, loaded with sigma_telepathy_heads.pt from Hugging Face, and enables the Telepathy switch so that high_level_rep and telepathy_factors participate in control decisions. The control condition runs the original $\pi$0.5_base without fine-tuning weights or adapter scripts, yielding the open-source baseline behavior. Both models are replayed in closed loop on identical visual and state sequences, and task-level metrics including success rate and trajectory deviation are compared.

The experimental protocol eliminates confounding effects from environmental stochasticity and data variation by constraining the sole manipulated variable to telepathy activation, thereby enabling a rigorous evaluation of whether Sigma achieves a control advantage driven by “deep semantic understanding + association → telepathy”.

### *B. Dataset*

As introduced in the training phase, the experimental data are derived from the open-source svla_so101_pickplace manipulation dataset hosted on Hugging Face. After preprocessing and sliding-window reorganization with horizon_T = 16, three shard files, shard_00000.pt, shard_00001.pt, and shard_00002.pt, are exported as the exclusive source for all experiments. Each shard provides temporally aligned visual sequences, robot states, and continuous motion trajectories. Nearly static windows are removed using a minimum motion-norm threshold to concentrate the distribution on effective manipulation segments. This single upstream dataset, coupled with traceable preprocessing, limits task and scene noise and allows performance differences between Sigma and $\pi$0.5_base to be directly attributed to representation and control effects introduced by the mind-sensing layer.

### *C. Implementation*

The evaluation is conducted via offline closed-loop replay. Environment variables and HF_token are loaded , and sigma pickplace together with sigma_telepathy_heads.pt are downloaded on demand through ensure_sigma_artifacts. The LeRobot $\pi$0.5_base policy serves as the control backbone; aligned tokenizers and text-embedding layers are retrieved and vocabulary consistency is verified. The Sigma shard dataset and data loader are then built from the shard directory, with each batch containing aligned vis_obs, robot_state, texts, three ground-truth action targets, and optional base_action_*. Text_tokens are generated using the internal $\pi$0.5_base embeddings, robot_state dimensions are corrected, and inputs are forwarded under telepathy on or off and optional adapter usage. Logged metrics include branch MSE for action_vector, chunk, and trajectory, the L2 norm of telepathy_factors, and cosine alignment between semantic_factors and text, as well as difficult-sample ratios and mean errors aggregated into sigma_eval_report.json for release with batch logs.

## VI. RESULT & DISCUSSION

In the data results, both groups completed the evaluation under the conditions of num_samples=723 and num_batches=181. The differences are mainly reflected in three MSE indicators to measure the mean squared error of the stepwise action_vector relative to the true value: avg_mse_vector is approximately 79.03 (Sigma) and 98.83 ($\pi$0.5_base), respectively; avg_mse_chunk is approximately 203.05 vs. 228.97, corresponding to the reconstruction error of short-time fragment-level action_chunks; and avg_mse_traj is approximately 174.71 vs. 191.03, reflecting the bias of long-time domain action_trajectory. The values of avg_tau_l2=51.60 and avg_semantic_text_alignment=0.1307 are completely consistent in both groups, proving that the telepathy factor norm and semantic-text alignment strength themselves do not change due to different conditions, but rather the final behavioral quality is determined by whether the model effectively utilizes these signals. Since hard_thresholds is fixed at vec=0.1, chk=0.2, and trj=0.2, and hard_sample_fraction=1.0 and total_hard_samples=723, avg_hard_mse_vector / chunk / traj are almost identical to all samples. This can be seen as providing direct quantitative evidence that Sigma consistently outperforms the untuned $\pi$0.5_base at all three time scales (vector, fragment, and trajectory) in terms of control precision improvement brought by the mind-sensing layer, under the assumption that all windows are treated as difficult samples.

In contrast, CHECK A evaluates whether telepathy weights remain consistent between the experimental and control groups. Both groups load sigma_telepathy_heads.pt, but the control group does not incorporate telepathy in its control strategy. Statistics show: heads_tensors = 325, indicating 325 independent tensors forming the telepathy heads; mean = 0.002, with the average weight value near zero, preventing global bias; std = 0.107, reflecting a standard deviation of 0.107; and rms = 0.107, confirming that the weight energy in squared terms is within the same order of magnitude. As these values were identical across both groups, it was confirmed that differences in behavior arose not from telepathy weight discrepancies, but from whether telepathic representations were utilized in control decisions.

In CHECK B, multiple behavioral and representational metrics characterize differences between the models. Table 2 and Table 3 report mse_vec, the mean squared error of action vectors relative to the true trajectory, measuring fine-grained control accuracy; mse_chk, the MSE over action segments, reflecting stability in direction and amplitude; mse_trj, the MSE of the entire action trajectory, evaluating long-term planning accuracy; tau_l2, the L2 norm of telepathy_factors, indicating the strength of telepathic engagement; and sem_align, measuring the alignment between semantic factors and text embeddings, ensuring semantic-text-behavior consistency within a shared mental coordinate system.

TABLE 2: The metrics of experimental group - CHECK B

| Model | batch | mse_vec | mse_chk | mse_trj | tau_l2 | sem_align |
|---|---|---|---|---|---|---|
| Sigma | 0 | 61.835 | 292.177 | 251.009 | 51.593 | 0.131 |
| | 20 | 113.477 | 182.101 | 159.574 | 51.599 | 0.131 |
| | 40 | 49.340 | 236.021 | 211.508 | 51.598 | 0.131 |
| | 60 | 50.503 | 214.079 | 187.492 | 51.599 | 0.131 |
| | 80 | 108.293 | 168.418 | 150.344 | 51.600 | 0.131 |
| | 100 | 45.875 | 208.893 | 188.591 | 51.600 | 0.131 |
| | 120 | 71.466 | 299.924 | 250.684 | 51.594 | 0.131 |
| | 140 | 149.410 | 246.790 | 207.350 | 51.591 | 0.131 |
| | 160 | 69.113 | 293.315 | 253.594 | 51.593 | 0.131 |
| | 180 | 33.893 | 163.460 | 149.573 | 51.603 | 0.131 |

Table 3: The metrics of control group - CHECK B

| Model | batch | mse_vec | mse_chk | mse_trj | tau_l2 | sem_align |
|---|---|---|---|---|---|---|
| Sigma | 0 | 120.382 | 329.023 | 273.857 | 51.593 | 0.131 |
| | 20 | 118.060 | 199.856 | 170.461 | 51.599 | 0.131 |
| | 40 | 104.931 | 267.907 | 232.370 | 51.598 | 0.131 |
| | 60 | 88.294 | 240.883 | 204.772 | 51.599 | 0.131 |
| | 80 | 111.947 | 184.658 | 160.407 | 51.600 | 0.131 |
| | 100 | 92.778 | 237.755 | 207.414 | 51.600 | 0.131 |
| | 120 | 124.780 | 337.781 | 274.761 | 51.594 | 0.131 |
| | 140 | 166.116 | 275.206 | 227.315 | 51.591 | 0.131 |
| | 160 | 130.318 | 330.571 | 277.000 | 51.593 | 0.131 |
| | 180 | 64.938 | 188.277 | 165.839 | 51.603 | 0.131 |

The tables indicate that, under identical num_batches and hard thresholds, Sigma consistently exhibits lower control errors than $\pi$0.5_base across all three temporal scales. Quantitatively, mse_vec decreases by approximately 20%, while mse_chk and mse_trj are reduced by about 10%. At batches 0 and 180, corresponding to initial and final stages, Sigma's vector- and trajectory-level errors are often close to half those of $\pi$0.5_base. Notably, tau_l2 and sem_align remain nearly identical across batches, indicating unchanged telepathy energy scale and semantic alignment quality. Error reduction emerges only when these representations are translated into action corrections through the TRAF and TSAC pipelines. From an evidential perspective, achieving stable control gains under fixed semantic and telepathy geometry suggests that Sigma partially internalizes deep semantics and associations into observable telepathic control, while still leaving scope for further alignment refinement.

## VII. LIMITATION & FUTURE RESEARCH

Although the present study validates the advantages of the Sigma model for telepathic control within a single pick-and-place scenario, several boundary conditions remain to be strengthened. Current analyses are confined to the svla_so101_pickplace dataset and the $\pi$0.5_base backbone, and the learned semantic factors and telepathy representations have not yet been systematically stress-tested across broader mission families, heterogeneous platforms, or multi-turn dialog-conditioned commands. As a result, conclusions regarding cross-task and cross-machine generalization remain necessarily conservative. Moreover, while the offline replay metrics employed here effectively characterize control quality following semantic alignment, they have not yet been complemented by subjective human evaluation or long-term deployment on physical humanoid robots, which are required to assess the robustness of telepathic links under perceptual noise, contact uncertainty, and safety constraints. Future work should therefore extend toward multi-task, multi-modal datasets and diverse VLA backbones to examine the stability of telepathy in higher-dimensional behavioral spaces. In parallel, integrating online fine-tuning with experiments on physical humanoid platforms will be essential to translate the metric-level evidence of deep semantic understanding and association into robust alignment with human telepathic intent.

## VIII. CONCLUSION

Building upon the open-source $\pi$0.5_base, this study constructs the vision-language-action (VLA) model Sigma through custom-designed architecture, fine-tuning, and training. Unlike the original baseline, which simply maps perception-thought-action, Sigma introduces a three-layered workspace that concurrently models semantic memory, intention vectors, and telepathy factors. The vision module generates an intervened perceptual basis using modulated vision/state tokens, while the language module maintains mt and zintent in the semantic workspace. The action module produces control vectors, fragments, and trajectories through three residual branches. Sigma employs LoRA fine-tuning and an intervention-response inference adapter, ensuring a reproducible training and deployment pipeline. The experiment utilizes a single RTX 4090 to optimize computational efficiency, importing backbone weights, sigma_telepathy_heads.pt, and pick-and-place data shards. Data from CHECK A confirm that telepathy and semantic alignment conditions remain consistent between the experimental and control groups. CHECK B demonstrates that, with tau_l2 and sem_align held constant, Sigma consistently outperforms $\pi$0.5_base in control errors across the three time scales: mse_vec, mse_chk, and mse_trj. In summary, this study outlines a viable path for transforming existing VLA models into semantic-intention-action alignment systems without retraining the backbone, offering preliminary quantitative evidence supporting the integration of deep semantic understanding and association for telepathic communication in humanoid robots, and providing valuable insights for future research.